\documentclass[conference]{IEEEtran}

\usepackage{cite}
\usepackage{amsmath,amssymb,amsfonts}
\usepackage{algorithmic}
\usepackage{graphicx}
\usepackage{textcomp}
\usepackage{xspace}
\usepackage{xcolor}
\usepackage{multirow}

\usepackage{array}
\newcolumntype{H}{>{\setbox0=\hbox\bgroup}c<{\egroup}@{}}
\newcolumntype{Z}{>{\setbox0=\hbox\bgroup}c<{\egroup}@{\hspace*{-\tabcolsep}}}

\usepackage{tikz}
\usetikzlibrary{shapes,arrows,chains,positioning}

\def\BibTeX{{\rm B\kern-.05em{\sc i\kern-.025em b}\kern-.08em
    T\kern-.1667em\lower.7ex\hbox{E}\kern-.125emX}}
\begin{document}

\title{A Scalable Handwritten Text Recognition System}

\newcommand{\todo}[1]{\textcolor{red}{TODO: #1}}
\newcommand{\todoa}[2]{\textcolor{red}{TODO(#1): #2}}
\newcommand{\eg}{e.g.\xspace}
\newcommand{\ie}{i.e.\xspace}
\newcommand{\bvec}[1]{\mbox{\boldmath $#1$}}

\author{
\IEEEauthorblockN{R. Reeve Ingle, Yasuhisa Fujii, Thomas Deselaers, Jonathan Baccash, Ashok C. Popat}
\IEEEauthorblockA{\textit{Google Research} \\
Mountain View, CA 94043, USA \\
Email: \{reeveingle,yasuhisaf,deselaers,jbaccash,popat\}@google.com}
}

\maketitle

\begin{abstract}

Many studies on (Offline) Handwritten Text Recognition (HTR) systems have focused on building state-of-the-art models for line recognition on small corpora. However, adding HTR capability to a large scale multilingual OCR system poses new challenges. This paper addresses three problems in building such systems: data, efficiency, and integration. 
Firstly, one of the biggest challenges is obtaining sufficient amounts of high quality training data. We address the problem by using online handwriting data collected for a large scale production online handwriting recognition system. We describe our image data generation pipeline and study how online data can be used to build HTR models. We show that the data improve the models significantly under the condition where only a small number of real images is available, which is usually the case for HTR models.  It enables us to support a new script at substantially lower cost. 
Secondly, we propose a line recognition model based on neural networks without recurrent connections. The model achieves a comparable accuracy with LSTM-based models while allowing for better parallelism in training and inference.
Finally, we present a simple way to integrate HTR models into an OCR system. These constitute a solution to bring HTR capability into a large scale OCR system.

\end{abstract}

\IEEEpeerreviewmaketitle

\section{Introduction}
\label{sec:introduction}

Handwritten text can be found in many types of images: handwritten notes, memos, whiteboards, medical records, historical documents, text input by stylus, etc. Therefore, a complete OCR solution has to include support for recognizing handwritten text in images.  This emphasizes the need for research into the area of building large scale handwriting recognition systems for many languages and scripts.

Many of the recent offline handwritten text recognition (HTR) systems have adopted line-level recognition strategies that use a combination of convolutional neural networks (CNN) and Long Short-Term Memory (LSTM)~\cite{hochreiter:lstm1997} recurrent neural networks for feature extraction, trained with Connectionst Temporal Classification (CTC~\cite{graves:icml2006})  loss~\cite{graves:nips2009,graves:pami2009,pham:icfhr2014,voigtlaender:icfhr2016,bluche:icdar2017,puigcerver:icdar2017,castro:icfhr2018}. 
Similarly to other fields, these data-driven deep-learning approaches learn features directly from the training data, in contrast to traditional methods that employ hand-engineered features. While these methods have resulted in substantial improvements in handwriting recognition accuracy on public benchmarks~\cite{pham:icfhr2014,voigtlaender:icfhr2016,puigcerver:icdar2017}, scaling these approaches to support new domains or new languages can be challenging due to the cost and difficulty associated with collecting and labeling a large corpus of handwritten training data of line images.

We believe that the biggest challenge in developing an HTR system is not modelling but obtaining sufficient amounts of high quality training data.   In this paper we address this problem by leveraging a large amount of online handwriting data and only few images of handwriting:
We use the \emph{ink} data from our online handwriting recognition system~\cite{Google:HWRPAMI,Google:HWRLSTM2018} to train the line recognizer in our existing OCR system~\cite{walker:das2018}. An \emph{ink} is the sequence of strokes of $(x,y)$ coordinates with timestamps of a user writing with their finger (or a stylus) on a screen. Online handwriting recognition is important in the context of entering text into computing devices in a world where large parts of humanity gain access to computing devices through mobile devices. Many of these users are using languages written in scripts that are not as easy to type as English~\cite{ghoshindickeyboardlayout}.

To the best of our knowledge, this is the first work that leverages the relationship between online and offline handwriting recognition by re-using an existing dataset of online handwriting samples to build a full HTR system.
Our HTR system is trained on images that are rendered from the trajectory data of the online handwriting recognition system.  To achieve sufficient accuracy on real images of handwriting, we have adopted an image degradation approach that is commonly used to generate realistic synthetic training images for OCR systems~\cite{tmb-tutorial-das2018}. 
After rendering clean handwriting images from trajectory data, the images are processed with an image degradation pipeline that applies realistic image transformations to the data. 
In addition to the collected online handwriting data we also use a handwriting synthesis pipeline to enrich the variability in our training dataset to obtain better results. 
We are demonstrating the feasibility of this approach on Latin script, but we believe that this approach will  work for other scripts as well. 

For the line recognizer we experiment with two different model architectures: We describe an LSTM-based architecture similar to many of the state-of-the-art methods. However, a problem with recurrent architectures is that they do not train and run as easily on specialized hardware (GPUs, TPUs) as feed forward networks. Thus we also propose a fully feed-forward network architecture that achieves comparable accuracy as our LSTM-based model.

While the line recognition engine is often the primary focus of HTR research, it is just one component of a full text recognition system. We describe the additional steps that were taken to integrate HTR support into an end-to-end text recgognition system consisting of text detection, direction identification, script identification, and text line recognition models~\cite{fujii:icdar2017}. 

The remainder of this paper is structured as follows: 

Section~\ref{sec:line_recognizer} describes our line recognition model. 
Section~\ref{sec:datapipeline} describes how we use online handwriting data to generate training images suitable for training an offline recognizer including the rendering process, the image degradation pipeline, and our handwriting synthesis system. In section~\ref{sec:multilingualhtrocr} we describe how the line recognizer is integrated into our full OCR system. Sections~\ref{sec:experimentalsetup} and \ref{sec:results} describe and discuss our experimental evaluation and section~\ref{sec:conclusion} concludes the paper.

\section{Handwritten Text Line Recognition}
\label{sec:line_recognizer}

\subsection{Model}

The task of handwritten text line recognition is to produce a sequence of Unicode code points for the handwritten text in a single line image. Let $\bvec{x}$ be a line image and $\bvec{y}$ be a sequence of Unicode points. We consider a probabilistic model $P(\bvec{y}|\bvec{x})$ to represent the relationship between $\bvec{x}$ and $\bvec{y}$. 

We follow the approach described in~\cite{fujii:icdar2017} to model $P(\bvec{y}|\bvec{x})$: We scale the height of the input image to a fixed size of $40$ pixels and collapse the color space down to luminance (gray-scale). The processed image is then fed into a neural network to produce a 1-D sequence of logits where each logit corresponds to a single character or a special blank symbol. The network is trained using a CTC loss~\cite{graves:icml2006}. The model naturally handles a variable width (length) of image. At inference time, the logits cost is combined with the cost of a character-based n-gram language model in a log-linear way and beam search is used to find the best recognition result. 

We experiment with two different model architectures:

\subsubsection{LSTM-based model}

LSTMs and other recurrent neural networks are dominantly used for modern handwritten text line recognition models~\cite{graves:nips2009, kozielski:icdar2013, doetsch:icfhr2014, pham:icfhr2014, voigtlaender:icassp2015, voigtlaender:icfhr2016, puigcerver:icdar2017, bluche:icdar2017}. Our model is inspired by the CLDNN (Convolutions, LSTMs, Deep Neural Network) architecture proposed by \cite{sainath2015convolutional}.
For the convolutional layers, we use the inception~\cite{szegedy:cvpr2015} style architecture described in~\cite{fujii:icdar2017}.
For the LSTM layers, we use between one and four stacked bidirectional LSTMs (BLSTMs). More details are given in Section~\ref{sec:experimentalsetup}.

\subsubsection{Recurrence-free model}
\label{sec:htrmodel}

\begin{figure*}[tbp]
\centering
\includegraphics[width=0.7\linewidth]{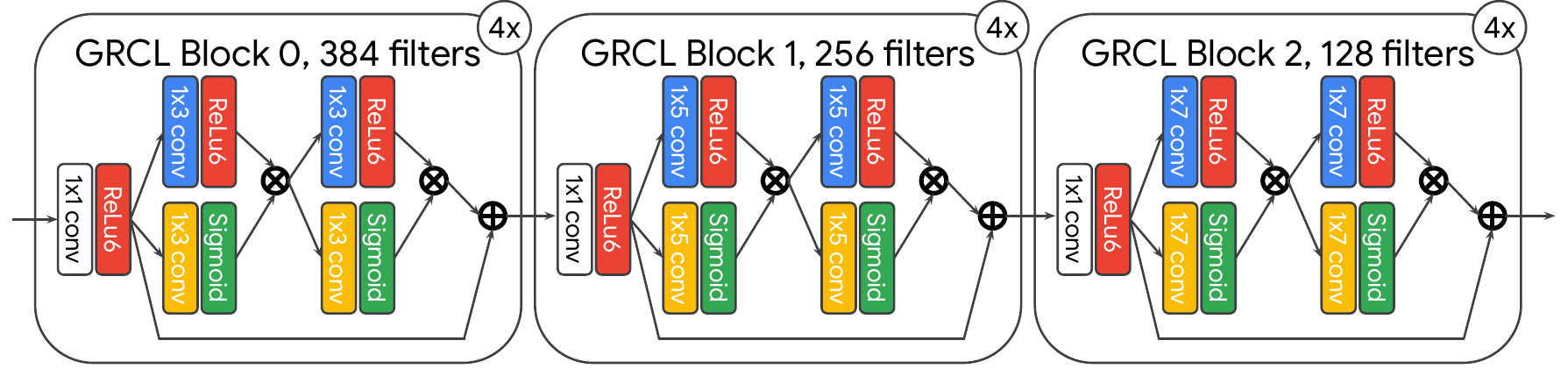}
\caption{Gated recurrent convolutional layer. Each GRCL block contains two convolutional layers with ReLU6 activations, with shared weights. The output of these layers are multiplied element-wise by the output of convolutional layers with sigmoid activations, which serve as a learned gating mechanism.  Convolutional weights for the gates are also shared within each GRCL block.}
\label{fig:gatedrcldecoder}
\end{figure*}

One disadvantage of recurrent models is that they are inherently sequential and cannot take full advantage of the parallelism offered by accelerators such as GPUs and TPUs. Therefore, network architectures having only feed-forward connections are preferred over those with recurrent connections for a large scale production system when comparable accuracy is achieved. In order to simulate the recurrency and gating mechanism of LSTMs with a feed-forward structure, we follow an idea from~\cite{wang:nips2017} and use 1-D gated recurrent convolutional layers (GRCLs)\footnote{The term "recurrent" in GRCL is used to mean recurrency along depth. The recurrency we want to avoid here is recurrency along time.} as a fully-feed-forward alternative to LSTMs (see Figure~\ref{fig:gatedrcldecoder}). Briefly speaking, GRCL is a recurrent convolutional network~\cite{liang:cvpr2015} with gating mechanism. The recurrency is along depth, and information propagation is controlled by the gating mechanism, which is computed using a sigmoid activation that is multiplied by the original ReLU activation. Although the recurrency is along depth, information can be propagated along time with a large receptive field. \cite{wang:nips2017} uses GRCLs as an alternative to CNNs, and the GRCLs are still followed by BLSTMs. We use GRCLs as an alternative to LSTMs, and the CNN layers are not changed. In this way, the model has only feed-forward connections. We show that the combination yields comparable accuracy with LSTM-based systems.\footnote{We can also consider CNN-only models to design feed-forward models as in~\cite{fujii:icdar2017}. However, we observed that such models performed significantly worse than LSTM-based models on handwritten text and did not explore those models further.}

\subsection{Training Data}

The model needs line images ($\bvec{x}$) along with their transcriptions ($\bvec{y}$) for training. One way to get such data is to manually annotate handwritten text lines in images collected from various sources. This provides the most valuable data. However, it is time-consuming and costly to do the annotation. Moreover, it is not trivial to find a sufficiently large amount of images containing handwritten text. Thus, the availability of manually labeled data tends to be limited especially when the goal is to support many languages.

In the following section we describe how we leverage a large amount of stroke data that was collected to build an online-handwriting recognition system~\cite{Google:HWRPAMI,Google:HWRLSTM2018}. This method enables us to support a new script at substantially lower cost.

\section{Data Synthesis, Rendering, and Degradation Pipeline}
\label{sec:datapipeline}

We use a mix of different sources of data to obtain the best possible HTR system: 
We have access to a \emph{large amount of ink data} collected for building an online handwriting recognition system \cite{Google:HWRPAMI,Google:HWRLSTM2018} in many languages. For the purpose of this paper we are only looking at Latin script but are planning to expand our system to other scripts and languages. 
We further are using an \emph{online handwriting synthesis pipeline} to enrich this data with handwriting styles that are not well represented in the original corpus. 
This data is rendered into images using a rendering pipeline described below and then degraded using the same degrading pipeline that we are using to train an OCR system on synthetic typeset data. To improve recognition accuracy on typeset text, the training data also includes degraded synthetic typeset data. Additionally, we include historic image data from several public datasets to improve recognition accuracy on historical handwriting~\cite{sanchez:icdar2015,toselli:icfhr2016,sanchez:icdar2017,emmo}.
Finally, we have a small amount of image data with modern handwriting labeled at the line level.

\begin{figure}
    \centering
    \newcommand{\ig}[1]{\includegraphics[height=0.8cm]{fig/real_data/#1}}
    \ig{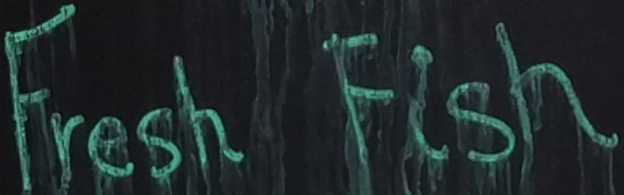}
    \ig{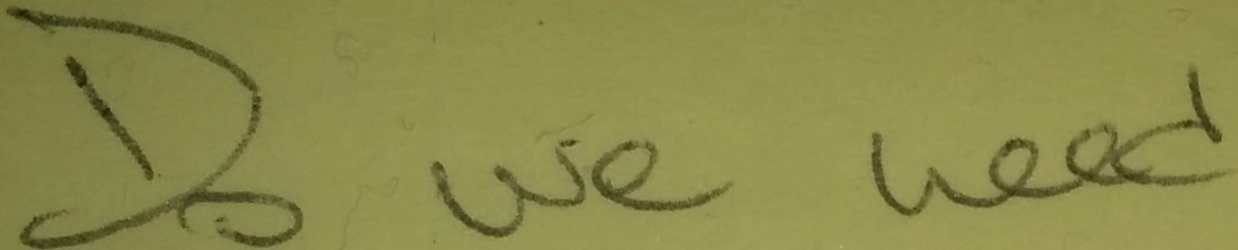}\\[1ex]
    \ig{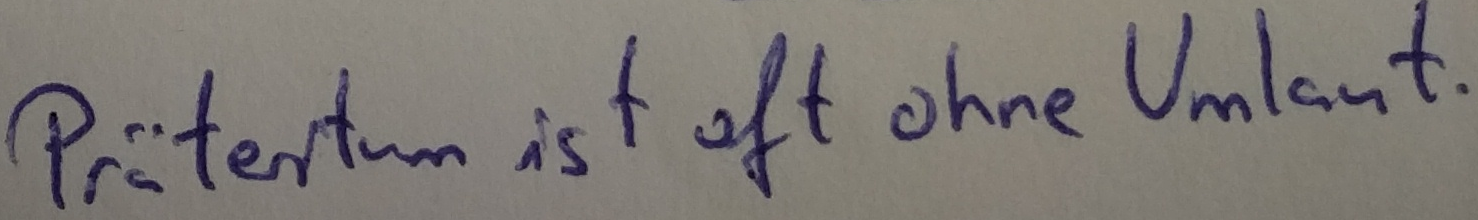}
    \ig{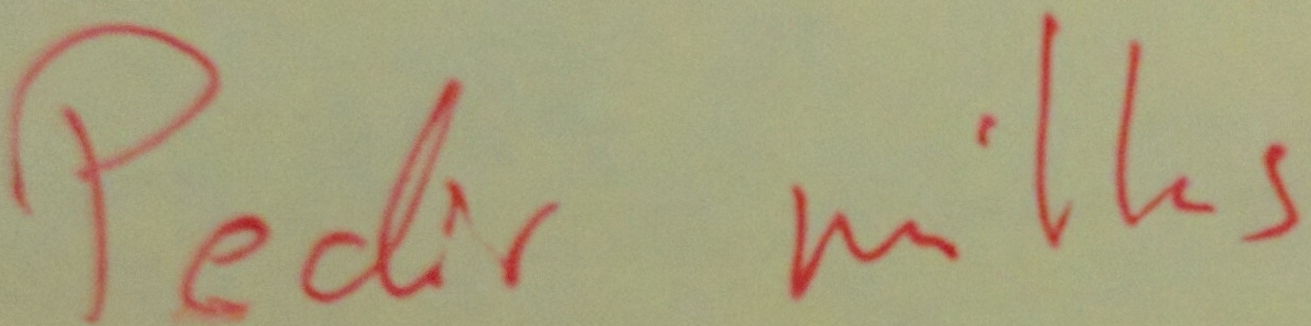}
    \caption{Example line images from our collected real handwriting image set. }
    \label{fig:real_data}
\end{figure}

Table~\ref{tab:dataset_sizes} gives an overview of the amount of data of each type that is used to train our final system and Figure~\ref{fig:example data} shows samples of our training data. Figure~\ref{fig:real_data} shows some samples of our labeled real image data. 

\begin{table}[tbp]
    \centering
    \caption{Number of training samples from different sources in our final system.}
    \label{tab:dataset_sizes}
    \begin{tabular}{|l|r|r|}
        \hline
        Type of data & \# training samples & \# test samples \\
        \hline
         Online data            & 1,115,822 & 34,895 \\
         Long lines data        &   500,000 &      0 \\
         Synthetic online data  & 1,033,680 &      0 \\
         Synthetic typeset data &   500,000 &      0 \\
         Labeled image data     &    62,477 &  8,362 \\
         Historic image data     &  159,326 &      0 \\
         \hline
    \end{tabular}
\end{table}

\begin{figure*}
    \centering

    \newcommand{\imgs}[2]{\includegraphics[height=.5cm]{fig/example-data/#1-rend-#2} & \includegraphics[height=.5cm]{fig/example-data/#1-degrade-#2}}
    \begin{tabular}{lll}
    \hline
    label & rendered & degraded\\
    \hline
    important & \imgs{orig}{01d7870240ed8789} \\
    because & \imgs{orig}{01d77e319ae1a0e1} \\
    down upon &\imgs{orig}{01d64473269f4608} \\
    \hline
    Zagrebu And fro & \imgs{long}{01d568f88ec5d3e1} \\
    and Two Viola Ground you & \imgs{long}{01c9f313a5d1bea6}\\
    \hline
    language of the law remains & \imgs{synth}{0006c54e135a890b}\\
    Academy of Design & \imgs{synth}{000751ea8a53a523}\\
    \hline
    PARADOXY A KRITIKA & \imgs{typeset}{47937965}\\
    o distribuie la microprocesor, & \imgs{typeset}{a132f504}\\
    \hline

    \end{tabular}
    \caption{Examples of label, rendered ink, and degraded image. The first 3 rows are samples that are rendered from the training data of the online handwriting recognition system (sec.~\ref{sec:rendering}). 
    The fourth and fifth rows are examples of a long line created by concatenating multiple short inks (sec.~\ref{sec:rendering}). The sixth and seventh rows are synthetic inks (sec.~\ref{sec:synthesis}). The final two rows are synthetic typeset text lines. The last column shows some typical degradations (sec.~\ref{sec:degrader}).}
    \label{fig:example data}
\end{figure*}

\subsection{Rendering Images from Stroke Data}
\label{sec:rendering}

The ink data that was used to train this system has been obtained through a variety of sources including prompted data collection of individual words on mobile phones and logged traffic of users of a handwriting recognition web service~\cite{Google:HWRPAMI,Google:HWRLSTM2018}. Most of the data was obtained with the goal of building a handwritten text input method for mobile phones and thus most of the written items are individual words, partial words, or even individual letters.  
In most cases the writing is in a single line, but in some cases  the data was written over multiple lines, \eg when users had no space to finish a word at the edge of the screen. 

To obtain data feasible to train an OCR line recognizer, we first processed the entire dataset using forced alignment to get character segmentation using our segment-and-decode online handwriting recognition  system~\cite{Google:HWRPAMI,Google:HWRLSTM2018}.
Then, we apply a simple heuristic to split multi-line inks into single lines: for each character, we look at the bounding box -- if the highest point of a new character is at most 20\% overlapping with the existing line, it is considered to be the start of a new line.  After this step all inks are single-line inks. 

Since most of the inks are short, we supplement these with longer lines that are created by randomly combining individual inks into longer lines. For this we concatenate the label (with a space) and arrange the corresponding inks with a randomly chosen amount of space in between~\cite{chen:icdar2017}. The target line lengths are distributed roughly uniformly between 500 and 2500, at a height of 100. 

Finally we render the inks into images using a randomly chosen stroke width, a randomly chosen slant, and a randomly chosen boundary size with black pixels on white background at an image height of 100 pixels. 
Figure~\ref{fig:example data} shows some examples of this data without and with concatenated long lines.

\subsection{Synthesis of Handwriting Stroke Data}
\label{sec:synthesis}

From inspecting our data, we observed  that the training data from the online handwriting recognition system has two deficiencies. 
1) Most of the written items are very short. We are addressing this by combining individual items into longer lines, however the resulting data, while long enough, does not create a realistic text sample, nor does it have a consistent writing style within the line.  
2) Most of the written items have well-separated characters (rather than a cursive style). This is due to most of the data coming from users entering text on mobile phones - where users write pragmatically. For handwriting recognition in images, we need to recognize a wider variety of styles, including print, cursive, and non-natural writing styles, \eg a restaurant menu written with neat, stylized handwriting. 

To alleviate these two problems we employ a \emph{handwriting synthesis pipeline}. This pipeline was trained on the same training data, but it allows for controlling the ``\emph{neatness}'' and the writing style. 

Our handwriting stroke synthesis pipeline is based on the system described in~\cite{graves:arxiv2013}. 
In addition to the original system, we have added a style embedding module to the network. The style embedding module consists of two LSTM layers that take as input the output from the original synthesis model. 
We compute the sum of the output of the style-embedding LSTM as a style embedding and pass that as an additional input to the mixture density output layer of the original synthesis model. This style embedding captures the writing style because input/output are independent of the label. The style embedding is trained as part of the normal training procedure without any extra loss function.

In order to synthesize handwriting with the style of a given sample, we compute the style embedding for that sample (by passing it through the style embedding directly) and then feed the fixed style vector into the mixture density layer when synthesizing. 
We use this method to synthesize additional cursive data by using the style embeddings from a small number of handpicked cursive writing samples. 

In order to obtain meaningful text samples, we synthesize the same text corpus as we use in the synthetic data pipeline of our OCR system. 

Finally, in order to create neat handwriting samples, we reduce the sampling temperature in the mixture density network layer - effectively sampling very close to the mean handwriting for a given style. This typically leads to visually appealing samples with a high regularity - as you expect when someone is trying to write neatly. 

See Figure~\ref{fig:example data} for some examples of synthetic handwriting samples.

\subsection{Image Degradation Pipeline}
\label{sec:degrader}

\begin{figure*}
\resizebox{\linewidth}{!}{
\begin{tikzpicture}[%
startstop/.style={rectangle, rounded corners, minimum width=1cm, minimum height=0.5cm,text centered, draw=black, fill=blue!30},
st/.style={rectangle, rounded corners, minimum width=1cm,text width=1cm, minimum height=0.5cm,text centered, draw=black, fill=blue!30},
conditional/.style={rectangle, rounded corners, minimum width=1cm,text width=1cm, minimum height=0.5cm,text centered, draw=black, fill=blue!30,rectangle split, rectangle split parts=2},
arrow/.style={thick,->,>=stealth}
]
\node (start) [startstop] {};

\node (scale_books) [st, right = 0.5cm of start] {scale};
\node (borderall)[conditional, right = 0.5cm of scale_books] {0.2\nodepart{second}border};
\node (border_line)[conditional, right = 0.5cm of  borderall] {0.1\nodepart{second}outline};
\node (border)[conditional, right = 0.5cm of  border_line] {0.3\nodepart{second}border};
\node (transform_weak)[conditional, right = 0.5cm of  border] {0.4\nodepart{second}transf.};
\node (offset)[conditional, right = 0.5cm of  transform_weak] {0.8\nodepart{second}offset};
\node (contrast)[conditional, right = 0.5cm of  offset] {0.8\nodepart{second}contrast};
\node (background)[conditional, right = 0.5cm of  contrast] {0.2\nodepart{second}BG};
\node (blur)[conditional, right = 0.5cm of  background] {0.8\nodepart{second}blur};
\node (noise)[conditional, right = 0.5cm of  blur] {0.8\nodepart{second}noise};
\node (gradient)[conditional, right = 0.5cm of  noise] {0.8\nodepart{second}gradient};
\node (quantization)[st, right = 0.5cm of  gradient] {quant.};
\node (invert)[conditional, right = 0.5cm of  quantization] {0.5\nodepart{second}invert};

\draw [arrow] (start) -- node[below] {0.7}  (scale_books);
\draw [arrow] (scale_books) -- (borderall);
\draw [arrow] (borderall) -- (border_line);
\draw [arrow] (border_line) -- (border);
\draw [arrow] (border) -- (transform_weak);
\draw [arrow] (transform_weak) -- (offset);
\draw [arrow] (offset) -- (contrast);
\draw [arrow] (contrast) -- (background);
\draw [arrow] (background) -- (blur);
\draw [arrow] (blur) -- (noise);
\draw [arrow] (noise) -- (gradient);
\draw [arrow] (gradient) -- (quantization);
\draw [arrow] (quantization) -- (invert);

\node (crop) [st, right = 0.5cm of start, below = 1.5cm of scale_books] {crop};
\node (scale_aspect) [st, right = 0.5cm  of crop] {AR};
\node (border_line_seq) [conditional, right = 0.5cm  of scale_aspect] {0.2\nodepart{second}border};
\node (border_photo) [conditional, right = 0.5cm  of border_line_seq] {0.8\nodepart{second}border};
\node (text_border) [conditional, right = 0.5cm  of border_photo] {0.1\nodepart{second}text border};
\node (text_color) [st, right = 0.5cm  of text_border] {color};
\node (fixheight60) [st, right = 0.5cm  of text_color] {H=60};
\node (transform_photo) [conditional, right = 0.5cm  of fixheight60] {0.8\nodepart{second}transf.};
\node (scale_photo) [st, right = 0.5cm  of transform_photo] {scale};
\node (background_photo) [st, right = 0.5cm  of scale_photo] {BG};
\node (fixheight60_2) [st, right = 0.5cm  of background_photo] {H=60};
\node (blur_photo) [conditional, right = 0.5cm  of fixheight60_2] {0.95\nodepart{second}blur};
\node (noise_photo) [st, right = 0.5cm  of blur_photo] {noise};
\node (gradient) [st, right = 0.5cm  of noise_photo] {grad.};
\node (quantization_photo) [conditional, right = 0.5cm  of gradient] {0.5\nodepart{second}quant.};
\node (jpeg) [conditional, right = 0.5cm  of quantization_photo] {0.2\nodepart{second}jpeg};
\node (invert) [conditional, right = 0.5cm  of jpeg] {0.5\nodepart{second}invert};
\node (fixheight40) [st, right = 0.5cm  of invert] {H=40};

\draw [arrow] (start) -- node[below] {0.2}  (crop);
\draw [arrow] (crop) -- (scale_aspect);
\draw [arrow] (scale_aspect) -- (border_line_seq);
\draw [arrow] (border_line_seq) -- (border_photo);
\draw [arrow] (border_photo) -- (text_border);
\draw [arrow] (text_border) -- (text_color);
\draw [arrow] (text_color) -- (fixheight60);
\draw [arrow] (fixheight60) -- (transform_photo);
\draw [arrow] (transform_photo) -- (scale_photo);
\draw [arrow] (scale_photo) -- (background_photo);
\draw [arrow] (background_photo) -- (fixheight60_2);
\draw [arrow] (fixheight60_2) -- (blur_photo);
\draw [arrow] (blur_photo) -- (noise_photo);
\draw [arrow] (noise_photo) -- (gradient);
\draw [arrow] (gradient) -- (quantization_photo);
\draw [arrow] (quantization_photo) -- (jpeg);
\draw [arrow] (jpeg) -- (invert);
\draw [arrow] (invert) -- (fixheight40);

\node (border) [conditional, right = 0.5cm of start, above = 1.5cm of scale_books] {0.5\nodepart{second}border};
\node (transform) [conditional, right = 0.5cm  of border] {0.5\nodepart{second}transf.};
\node (text_color) [conditional, right = 0.5cm  of transform] {0.5\nodepart{second}text color};
\node (background_bd) [conditional, right = 0.5cm  of text_color] {0.5\nodepart{second}BG bd};
\node (noise) [conditional, right = 0.5cm  of background_bd] {0.5\nodepart{second}noise};
\node (invert) [conditional, right = 0.5cm  of noise] {0.5\nodepart{second}invert};
\node (scaleall) [st, right = 0.5cm  of invert] {scale small};

\draw [arrow] (start) -- node[above] {0.1}  (border);
\draw [arrow] (border) -- (transform);
\draw [arrow] (transform) -- (text_color);
\draw [arrow] (text_color) -- (background_bd);
\draw [arrow] (background_bd) -- (noise);
\draw [arrow] (noise) -- (invert);
\draw [arrow] (invert) -- (scaleall);
\end{tikzpicture}}
\caption{A flow diagram describing the degradation system. Each box is a degradation step, numbers on arrows indicate probabilities that a path is chosen, numbers in the top part of a degradation step indicate probability that the node step is executed. 
Degradation steps are: 
border: adds a border to an image;
outline: adds a line around an image;
transf.: performs a  spline, rotation, or projective transformation;
BG: puts the text onto a background image;
scale: scales the image between 50\% and 150\%;
offset: applies an offset to the intensity values of all pixels in the image;
contrast: changes the contrast of the image;
blur: applies a blur filter to the image;
noise: adds additive noise to the image;
grad.: puts a gradient background behind the text;
quant.: quantizes the intensities in the image;
invert: inverts the image;
crop: crop up to 20\% of the image;
AR: randomly scales - changing aspect ratio;
text color: changes the intensity of the text with respect to the background;
jpeg: adds jpeg compression artifacts to the image;
H=X: scales the image to be of height X pixels;
text border: adds text fragments to the border of the image - mimicing bad segmentation.
}
\label{fig:degraders}
\end{figure*}
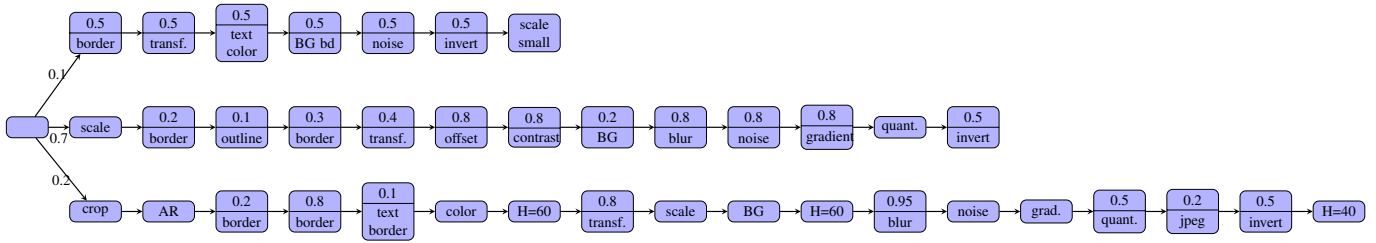

In order to make the images of rendered ink appear realistic, we apply our image degradation pipeline that we also use to degrade synthetic OCR training data. The degradation pipeline is a multi-step system where different degradation steps are performed in order. Most degradation steps use randomness to vary how they transform the input image, e.g. a ``background'' degrader has access to a large set of images, picks one of these at random and then superimposes the ink image onto this image, and a ``transform'' degrader would pick a transformation (e.g.\ scale, rotate) at random and then pick the parameters for the transformation at random. The steps of our degradation pipeline are shown in Figure~\ref{fig:degraders}. Figure~\ref{fig:example data} shows some images before and after the degradation process.

\section{Integration into the OCR system}
\label{sec:multilingualhtrocr}

\begin{figure*}[!t]
\centering
\includegraphics[width=0.8\linewidth]{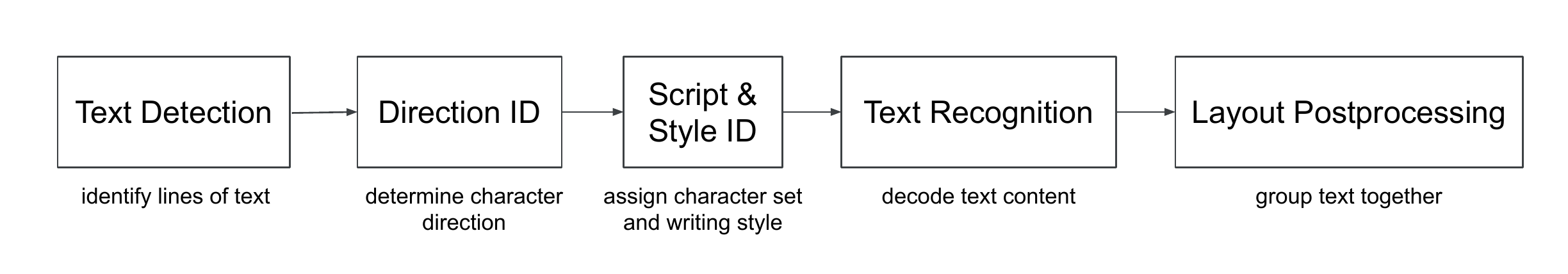}
\caption{Multilingual HTR and OCR system with line-level script identification and writing style classification.}
\label{fig:ocrsystem}
\end{figure*}

Line recognition is a component of a full text recognition system. We briefly describe our line-based OCR system and propose a simple way to integrate the handwritten text line recognition into the full system. 

\subsection{Line-based Multilingual OCR System}
\label{sec:ocrsystemoverview}

Figure~\ref{fig:ocrsystem} shows the end-to-end pipeline of our line-based multilingual OCR system.  The first step extracts text lines from the image. Next, direction identification determines the writing direction of each line and rotates the line images to make the characters in the images upright. Once the directions are fixed, script identification classifies the lines into scripts. Text recognition selects a model based on the script identification result and runs the model to obtain a sequence of Unicode points for each line. Finally, layout postprocessing infers the structure of the text (e.g. paragraph). The recognition results may be enhanced during the process, for example, by filtering results with low confidence values and by removing overlapping lines. Up to the layout postprocessing step, lines are the fundamental processing unit. More information on the system can be found in \cite{walker:das2018}.

\subsection{Dual-Head Script and Style Identification}
\label{sec:dualheadscriptid}

\begin{figure}[tbp]
\centering
\includegraphics[width=\linewidth]{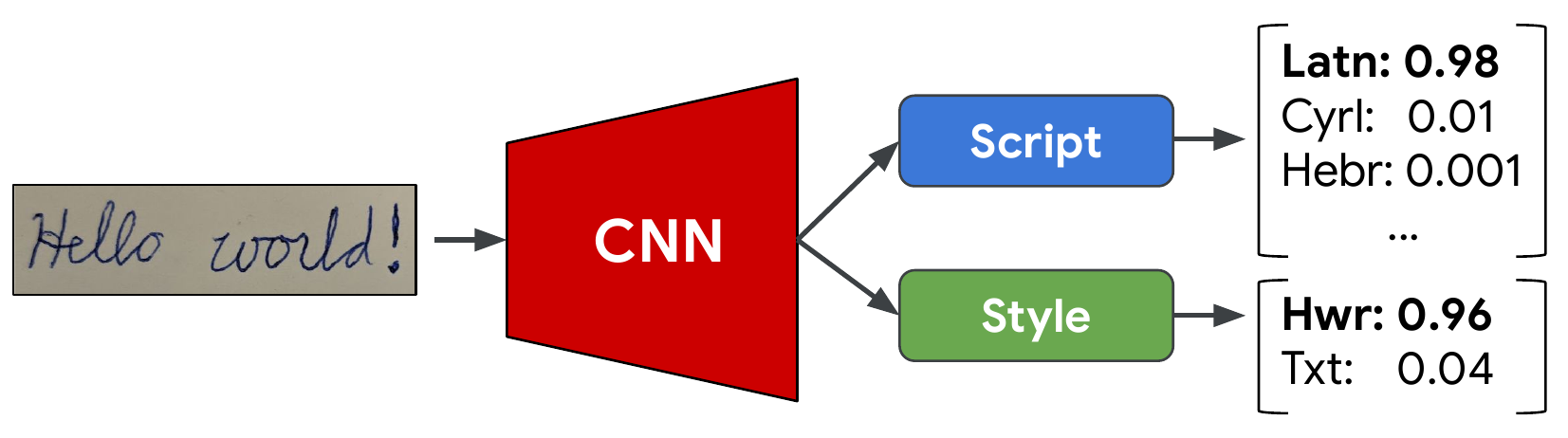}
\caption{Dual-head script and style classifier.}
\label{fig:dualheadscriptid}
\end{figure}

Script identification is the third step in our pipeline and is responsible for selecting the appropriate recognition model for each line. In order to integrate the handwriting line recognizer, we extended our model to allow for identifying a line as handwritten or printed which then selects the appropriate model.

We adopt the \emph{sequence-to-label} paradigm proposed in~\cite{fujii:icdar2017} to implement the script identification model. The model performs a single classification for the identification. We modify the model by adding an extra classification head to perform the style classification between printed and handwritten text. This approach is more effective than the naive approach where each script has has a `printed' and a `handwritten' sub-class, which doubles the number of classes in the model.

Figure~\ref{fig:dualheadscriptid} shows the proposed dual-head script and style identification model. The script identification head selects a script and the style identification head selects a style. The handwritten text model is selected if the probability of handwritten text is greater than or equal to a predefined threshold $\theta^{\text{hwr}}$.

\section{Experimental Setup}
\label{sec:experimentalsetup}

\subsection{Data}
\label{sec:data}

We have been using data from three different datasets for our experiments: 
The IAM offline handwriting database, the IAM online handwriting database, and an internal dataset of both online and offline handwriting samples. 

The IAM offline handwriting database~\cite{marti:ijdar2002} (IAM-Off) contains scanned pages of handwritten text passages, which were written by 500 different writers using prompts from the Lancaster-Oslo/Bergen (LOB) text corpus.  The database contains line images that have been partitioned into writer-disjunct training, validation, and test sets containing 6161, 976, and 2915 lines, respectively.

The IAM online handwriting database~\cite{liwicki:icdar2005} (IAM-On) contains trajectory data of handwritten text written by 221 different writers, acquired on a digital whiteboard. The database is partitioned into training, validation, and test sets containing 5363, 1438, and 3859 lines, respectively. Text prompts for the IAM online handwriting database were also drawn from the LOB corpus. In contrast to the IAM offline database, the partitioning into train, validation, and test sets did not ensure that the train/valid/test sets use a disjunct set of prompts. For our experiments, we have rendered the IAM online database using the method described in section~\ref{sec:rendering}. 
We also created an augmented version of this dataset (IAM-On-Long) containing 500'000 longer lines for training and 20'000 lines for validation by randomly concatenating samples as described in section~\ref{sec:rendering}

Our internal dataset consists of a large dataset of online handwriting samples \cite{Google:HWRPAMI,Google:HWRLSTM2018}, long lines obtained by concatenating samples from this dataset (see sec.~\ref{sec:rendering}), and synthetic samples generated from a model trained on this dataset (see sec.~\ref{sec:synthesis}). In addition we have a smaller set of scanned and photographed handwriting images for training and evaluation. The sizes of these individual datases are given in Table~\ref{tab:dataset_sizes} and Figure~\ref{fig:example data} shows some samples for each of these datasets. 

For the full system experiments, two datasets are used: printed text images and handwritten text images. The printed text images come from the Web and consist of 508 images with 4868 lines. The handwritten text images are a subset of the internal dataset, explained above, having full image annotations. The internal dataset contains some data that consists solely of line images, and these are excluded from the full system experiments. The data consists of 433 images with 2555 lines.

\subsection{Models}
\label{sec:models}

All neural network models were implemented with TensorFlow~\cite{TensorFlow:AbadiMartin:2016:OSDI}. A standard asynchronous stochastic gradient descent with multiple workers was used to train all models. The learning rate was decayed exponentially through the training steps. The hyper parameters of the training were tuned on the development set if available; otherwise on a holdout set of the training data. 

We trained two types of handwritten text recognition models. One uses BLSTM layers and the other uses GRCL blocks described in Section~\ref{sec:htrmodel}. For the former, we trained models with 1, 2, and 4 BLSTM layers. For the latter, we trained a model with the topology shown in Figure~\ref{fig:gatedrcldecoder}. The full model architecture is given in Appendix~\ref{sec:line-recognition-model-topology}. The models were combined with a character-based $n$-gram language model trained on text collected from the Web and Wikipedia. We used $n=9$.

We implemented the dual-head model for script and style classifications by adding an extra summarizer to the shared inception-style encoder. The full model architecture is given in Appendix~\ref{sec:dual-head-topology}. We incorporated the handwritten recognition model into the multilingual OCR system described in\cite{fujii:icdar2017} by using the dual-head script and style identifier. We treated the data used for~\cite{fujii:icdar2017} as the printed text data to train the style identifier. The OCR (printed text) model was trained with the printed text data used in~\cite{fujii:icdar2017} with a small amount of additional handwritten text data. Thus, the OCR model should be able to recognize handwritten text to some extent. However, as we observe in the experimental results, the handwritten-specific model is superior to the printed text model on handwritten text. We expect that the dual-head identifier provides better results than either of these models in isolation.

\subsection{Evaluation}
\label{sec:evaluation}

For line recognition experiments, we use character error rate (CER) and word error rate (WER) as error metrics. CER is defined as a ratio of the edit distance between the target and the reference character sequences to the number of characters in the reference. WER is defined analogously on words.

For the full system experiments, we use word recall and precision as error metrics. In order to consider the positional information in the metrics, we compute IoU between word pairs and ignore ones whose IoU are less than 0.01. We use the small number for the threshold to make the score less sensitive to the errors of bounding boxes in groundtruth. If a word in the recognition result does not overlap with any words in the reference, the word is considered as a false positive. If a word in the reference does not match with any words in the recognition result, the word is considered as a false negative. AUC (Area Under a Curve) is computed with recall and precision by changing the threshold to remove words whose confidence value is lower than the threshold.

\section{Results and Discussions}
\label{sec:results}

\subsection{Line Recognition}

\begin{table}[tbp]
\caption{Line-level results of the GRCL and two-layer BLSTM system on the IAM-Offline test data using different training sets.}
\label{tab:iam-results}
\resizebox{\linewidth}{!}{%
\begin{tabular}{|lr|rrHZ|rrHZ|}
\hline
 & & \multicolumn{2}{c}{GRCL}&& & \multicolumn{2}{c}{BLSTM}&&\\
\hline
 Training Data & \# tr.\ obs.& WER & CER & WER & CER & WER & CER & WER & CER\\
\hline
IAM-Off                  &   6'161& 35.2 & 14.1 & 57.0 & 31.4  & 30.7 & 12.8 & 53.6 & 29.2 \\
IAM-On                   &   5'363& 27.8 & 12.0 & 22.0 & 10.0  & 21.7 &  9.1 & 16.3 &  6.8 \\
IAM-On-Long              & 505'363& 24.8 & 10.5 & 20.5 &  9.0  & 18.3 &  7.4 & 14.9 &  5.9 \\
IAM-Off + IAM-On-Long-ND & 511'524& 22.3 &  8.8 & 22.4 &  9.4  & 23.0 &  9.2 & 21.4 &  8.8 \\
IAM-Off + IAM-On-Long    & 511'524& 17.0 &  6.7 & 18.2 &  7.9  & 15.1 &  5.7 & 14.2 &  5.6 \\
\hline
\end{tabular}
}
\end{table}

Table~\ref{tab:iam-results} shows line-level WER and CER of our GRCL-based and our LSTM-based system on the IAM offline  test data set  while varying the training data sets. As training data we used the IAM Offline training dataset, the IAM online training dataset with and without long-lines, rendered as described in section~\ref{sec:rendering} and also with no degradations (ND), as well as a combination of these. No additional sources of training data were used when training these models. 

The most surprising finding was that using the online dataset for training we obtain better results than when using the offline training data. The reason for this is that the \emph{effective} size of the online training set is larger than the 5'363 samples because we add random degradations during training (see section~\ref{sec:degrader}), which we do not do on the natural images in the IAM offline dataset. These random degradations facilitate training and help to mitigate overfitting.
Further, we found that it helps substantially to add the concatenated long lines into the training set, and that a combination of both datasets, with random degradations applied to the IAM online data, leads to the best results. This trend is identical for both the GRCL and the LSTM model.

\begin{table}[tbp]
\caption{Line-level results of our system trained on internal data and evaluated on internal data and the IAM offline testdata.}
\centering
\begin{tabular}{|lr|r r|r r|}
\hline
 & & \multicolumn{2}{c|}{\textbf{Off-Test}} & \multicolumn{2}{c|}{\textbf{Internal}}\\ 
\hline
 Model   &\#params& WER & CER & WER & CER \\
\hline
1 BLSTM             & 3'913'065 &  16.4 & 6.1 & 32.5 & 13.4 \\
2 BLSTM             & 4'307'305  &  14.9 & 5.5 & \bf 30.1 & \bf 12.4 \\
4 BLSTM             & 5'095'785  &  15.7 & 5.8 & 32.1 & 13.1 \\
GRCL                & 10'614'889  &  14.1 & 5.4 & 30.8 & 12.6 \\
GRCL + tuning      & &  \bf 10.8 & \bf 4.0 & 38.6 & 15.4 \\
\hline
Bluche et al.~\cite{bluche:icdar2017} && 10.5& 3.2 \\% &--& -- \\
Voigtlander et al.~\cite{voigtlaender:icfhr2016}& & 9.3 &3.5 \\ %
Poznanski et al.~\cite{Poznanski2016CNNNGramFH}& & 6.4 & 3.4\\ %
Castro et al.~\cite{castro:icfhr2018} && 10.5 & 3.6\\%&--& -- \\
\cline{1-4}
\end{tabular}
\label{tab:internal-results}
\end{table}

Table~\ref{tab:internal-results} shows results on the IAM Offline test data and on our internal dataset for 
a few different variants of our full line recognition system trained on internal data.
The first four rows were trained on the entire training data as described in Table~\ref{tab:dataset_sizes} and the row ``GRCL + tuning'' is the same system as the GRCL system but tuned to perform well on the IAM dataset by 
1) adding the IAM offline training data 50 times to the training data;
2) using a language model trained on the LOB corpus;
3) tuning the decoder weights on the IAM validation set.

For the internal dataset, we observe that the LSTM model with 2 layers performs best, larger models become worse, probably because of overfitting. On the IAM test set, the GRCL model performs a bit better than the best LSTM model, probably also because of less overfitting. This is surprising given that the number of parameters of GRCL model is as twice as the biggest BLSTM model (10'614'889 vs. 5'095'785). GRCL is less prone to overfitting than LSTM. Tuning the GRCL model for the IAM dataset leads to a substantial improvement of 25\% relative from 5.4\% to 4.0\% CER - which is close to the state of the art.

\begin{table}[tbp]
    \caption{Experiments evaluating the impact of different training datasets on the recognition accuracy using a GRCL-based system (compare line GRCL in table~\ref{tab:internal-results}). 
    The columns indicate which of the training datasets were used: 
    (1) labeled image data, 
    (2) rendered online data including long lines data, 
    (3) synthetic typeset data, 
    (4) synthetic online data, 
    (5) historic data. }
    \label{tab:data_ablation}
    \newcommand{\x}{+}
    \centering
    \begin{tabular}{|Z|c|c|c|c|c|cZH|r|r|r|r|}
\hline
         \multicolumn{9}{|c|}{Training dataset used} & \multicolumn{2}{c|}{\textbf{IAM-Off-Test}} & \multicolumn{2}{c|}{\textbf{Internal}}\\
\hline
hidden run name                                           &   &(1) &(2) &(3) &(4) &(5) &(6) &(7) & WER   & CER   & WER   & CER   \\ 
\hline

offhwr-mul-Latn-20190212-prod-realoffhwr-only             & A & \x &    &    &    &    &    &    & 45.7 & 19.5 & 58.8 & 27.2 \\ 
offhwr-mul-Latn-20190212-prod-online-render-only          & B &    & \x &    &    &    &    &    & 19.2 & 7.0  & 42.1 & 18.6 \\ 
offhwr-mul-Latn-20190212-prod-synthtypesetonly-fixdev     & C &    &    & \x &    &    &    &    & 51.8 & 25.4 & 54.6 & 27.0 \\ 
								          
offhwr-mul-Latn-20190212-prod-online-render-synth         & D &    & \x &    & \x &    &    &    & 20.9 & 7.7  & 42.5 & 18.3 \\ 

offhwr-mul-Latn-20190212-prod-onlinerendsynth-realoffhwr  & E & \x & \x &    & \x &    &    &    & 18.5 & 6.6  & 36.5 & 15.1 \\ 
offhwr-mul-Latn-20190212-prod-onlinerendsynth-synthtypeset& F &    & \x & \x & \x &    &    &    & 18.6 & 6.8  & 35.8 & 14.7 \\ 
								          
offhwr-mul-Latn-20190212-prod-nohist                      & G & \x & \x & \x & \x &    &    &    & 17.5 & 6.3  & 33.7 & 13.7 \\ 
offhwr-mul-Latn-20181213-rendered-20181108-iamalt         & H & \x & \x & \x & \x & \x &    &    & 14.1 & 5.4  & 30.8 & 12.6 \\ 
\hline
    \end{tabular}
\end{table}

Table~\ref{tab:data_ablation} shows the impact of using the different parts of our training datasets on the error rates on the IAM Offline test data and on our internal dataset.
The first interesting result is the difference between lines A and B where we found that using the large dataset of rendered online handwriting data (line B) leads  to much better results than training on a (much smaller) dataset of real handwriting images (line A). When training the recognizer only on (synthetic) printed OCR training data, results are very poor (line C).

Another interesting insight is that adding the dataset of rendered synthetic handwriting data to the handwriting data alone does not lead to an improvement but rather to a slightly worse result (line D). 
Adding the real handwriting data (line E) or the synthetic printed OCR data (line F) to this experiment gives a clear improvement in both cases and very similar results. 
Finally, line G, shows the yet again improved result obtained from combining all of the previous data sources and line (H) corresponds to additionally using historic data as training data - which leads to a surprisingly large improvement on both the IAM Offline test data and on our internal dataset although neither of these datasets contain any historic samples.

\subsection{Integration to the OCR System}

Table~\ref{tab:full-system-results} shows results for the full text recognition system with and without the dual-head script and style identifier. We used $\theta^{\text{hwr}}=0.95$ in the experiments. As expected, we observe that the OCR model outperforms the HTR model on printed text while the HTR model outperforms the OCR model on handwritten text. The dual-head system yields comparable accuracy with the OCR model and the HTR model on printed text and handwritten text, respectively. 98.9\% of the lines from the printed text images were classified as printed text by the style identifier, and 59.6\% of the lines from the handwritten text images were classified as handwritten. The low handwriting classification accuracy is due to the high classification threshold ($\theta^{\text{hwr}}=0.95$) as well as the fact that handwritten images frequently contain some lines of printed text. 

\begin{table}[tbp]
    \centering
    \caption{Word precision and recall of full text recognition system.}
    \begin{tabular}{|c|ccc|ccc|} \hline
        & \multicolumn{3}{c|}{Printed} & \multicolumn{3}{c|}{Handwritten} \\ \hline
        Model & Prec. & Recall & AUC & Prec. & Recall & AUC \\ \hline
        OCR & \bf 0.705 & \bf 0.702 & 0.594 & 0.562 & 0.596 & 0.449 \\
        HTR & 0.678 & 0.645 & 0.562 & 0.584 & 0.602 & \bf 0.476 \\
        Dual-head & \bf 0.705 & 0.701 & \bf 0.595 & 0\bf .586 & \bf 0.612 & 0.474 \\ \hline
    \end{tabular}
    \label{tab:full-system-results}
\end{table}

\section{Conclusion}
\label{sec:conclusion}

We described three techniques to build a scalable handwritten text recognition system.

We explored the use of online handwriting data to train handwritten text recognition models. We described the data generation pipeline and presented a series of techniques to generate better data. The experimental results showed that models can be improved by using the generated data, and we obtained the best model by combining the generated data and small amount of real images. 

When a large mount of data is available, the model with GRCL blocks was comparable with the model with LSTM layers. GRCL blocks are a good alternative to LSTM layers in a situation where feed-forward networks are preferred over recurrent networks for performance reasons.

We integrated the handwritten recognition model into the full text recognition system by augmenting the script identification model with an additional classification between printed text and handwritten text. The dual-head model yielded the best accuracy on both printed and handwritten text compared to printed and handwritten-specific models.

We applied these techniques only to Latin script. It is future work to apply them to other scripts and confirm the effectiveness of the approach on more scripts.

\section*{Acknowledgement}
We would like to thank Andrii Maksai and Pedro Gonnet for their work on the stroke synthesis pipeline, Henry Rowley and Philippe Gervais for their contributions to the rendering pipeline, Joan Puigcerver for helpful discussions during the preparation of this manuscript, and the entire Google OCR team for their contributions.

\bibliographystyle{IEEEtran}
\bibliography{IEEEabrv,main}

\appendix

\subsection{Line Recognition Model}
\label{sec:line-recognition-model-topology}

\begin{figure}[h]

\centering
\includegraphics[width=0.45\linewidth]{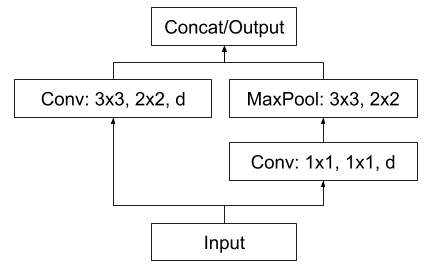}
\includegraphics[width=0.45\linewidth]{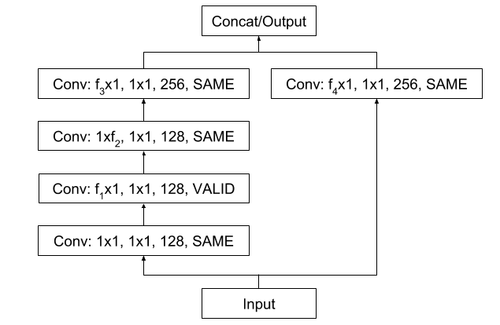}
\caption{Left: Pool module: ``Conv: $h_f \times w_f, h_s \times w_s, d$'' represents a convolutional neural network layer of filter size $h_f \times w_f$, stride $h_s \times w_s$, channels $d$, ReLU6 activation, and SAME padding. ``MaxPool: $h_f \times w_f$'' represents a max pooling layer of filter size $h_f \times w_f$, stride $1 \times 1$, and SAME padding.}
\label{fig:pool-module}
\caption{Right: Reduce module: ``Conv: $h_f \times w_f, h_s \times w_s, d, p$'' represents a convolutional neural network layer of filter size $h_f \times w_f$, stride $h_s \times w_s$, channels $d$, ReLU6 activation, and padding $p$.}
\label{fig:reduce-module}
\end{figure}

\begin{figure}[h]
    \centering
    \includegraphics[width=0.45\linewidth]{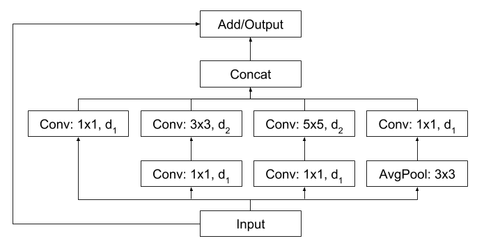}
\caption{Inception module: ``Conv: $h_f \times w_f, d$'' represents a convolutional neural network layer of filter size $h_f \times w_f$, stride $1 \times 1$, channels $d$, ReLU6 activation, and SAME padding. ``AvgPool: $h_f \times w_f$'' represents an average pooling layer of filter size $h_f \times w_f$, stride $1 \times 1$, and SAME padding.}
\label{fig:inception-module}
\end{figure}

Figure~\ref{fig:encoder} shows our inception-style CNN used for all line recognition models. 
LSTM-based models use stacked BLSTMs on top of this network. 
GRCL models use the GRCL layers shown in Figure~\ref{fig:gatedrcldecoder} on top of the network. Figures~\ref{fig:inception-module}, \ref{fig:pool-module}, and \ref{fig:reduce-module} give more details on the building blocks.

\subsection{Dual-head Script and Style Identifier Model}
\label{sec:dual-head-topology}

\begin{figure}[h]
\centering
\includegraphics[width=0.38\linewidth]{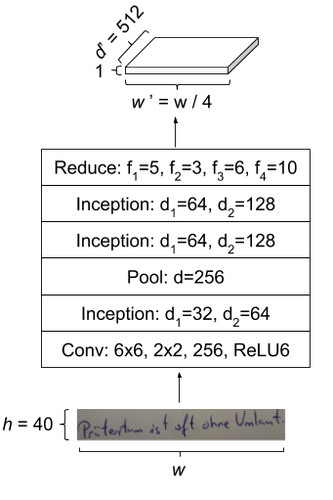}
\includegraphics[width=0.3\linewidth]{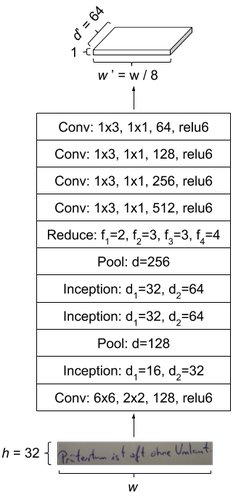}
\caption{Left: Inception-style CNNs used for all line recognizers: ``Conv: $h_f \times w_f, h_s \times w_s, d, f$'' represents a convolutional neural network layer of filter size $h_f \times w_f$, stride $h_s \times w_s$, channels $d$, activation $f$, and SAME padding. The Inception module, the Pool module, and the Reduce module are shown in Figure~\ref{fig:inception-module}, Figure~\ref{fig:pool-module}, and Figure~\ref{fig:reduce-module}, respectively.}
\caption{Right: Encoder used for the dual-head script and style identifier.}
\label{fig:encoder}
\label{fig:dual-head-encoder}
\end{figure}

Figure~\ref{fig:dual-head-encoder} shows the encoder network used for the dual-head script and style identifier. The output of the encoder network is fed to two summarizer networks for script and style prediction. We use the same Gate summarizer described in~\cite{fujii:icdar2017}.

\end{document}